\documentclass{article}

\usepackage{arxiv}

\usepackage{amsmath}
\usepackage{booktabs}
\usepackage{multirow}

\usepackage[utf8]{inputenc} 
\usepackage[T1]{fontenc}    
\usepackage{hyperref}       
\usepackage{url}            
\usepackage{booktabs}       
\usepackage{amsfonts}       
\usepackage{nicefrac}       
\usepackage{microtype}      
\usepackage{cleveref}       
\usepackage{lipsum}         
\usepackage{graphicx}
\usepackage{natbib}
\usepackage{doi}

\title{Controlling the Quality of Distillation in Response-Based Network Compression}

\author{ {Vibhas Vats} \\
	Indiana University Bloomington\\
	\texttt{vkvats@iu.edu} \\
	\And
	{David Crandall} \\
	Indiana University Bloomington\\
	\texttt{djcran@indiana.edu} \\
}

\renewcommand{\shorttitle}{\textit{arXiv} Template}

\hypersetup{
pdftitle={Controlling the Quality of Distillation in Response-Based Network Compression},
pdfsubject={q-bio.NC, q-bio.QM},
pdfauthor={Vibhas Vats, David Crandall},
pdfkeywords={Knowledge Distillation, Distillation, Distillation Hypothesis, Response-based Network Compression},
}

\begin{document}
\maketitle

\begin{abstract}
	The performance of a distillation-based compressed network is governed by the quality of distillation. The reason for the suboptimal distillation of a large network (teacher) to a smaller network (student) is largely attributed to the gap in the learning capacities of given teacher-student pair. While it is hard to distill all the knowledge of a teacher, the quality of distillation can be controlled to a large extent to achieve better performance. Our experiments show that the quality of distillation is largely governed by the quality of teacher's response, which in turn is heavily affected by the presence of similarity information in its response. A well-trained large capacity teacher loses similarity information between classes in the process of learning fine-grained discriminative properties for classification. The absence of similarity information causes the distillation process to be reduced from \textit{one example-many class} learning to \textit{one example-one class} learning, thereby throttling the flow of diverse knowledge from the teacher. With implicit assumption that only the instilled knowledge can be distilled, instead of focusing only on the knowledge distilling process, we scrutinize the knowledge inculcation process. We argue that for a given teacher-student pair, the quality of distillation can be improved by finding the sweet spot between batch size and number of epochs while training the teacher. We discuss the steps to find this sweet spot for better distillation. We also propose \textit{the distillation hypothesis} to differentiate the behavior of the distillation process between knowledge distillation and regularization effect. We conduct all our experiments on three different datasets.
\end{abstract}


\section{Introduction}\label{Introduction}

\begin{figure}[t]
    \centering
    \includegraphics[scale=1]{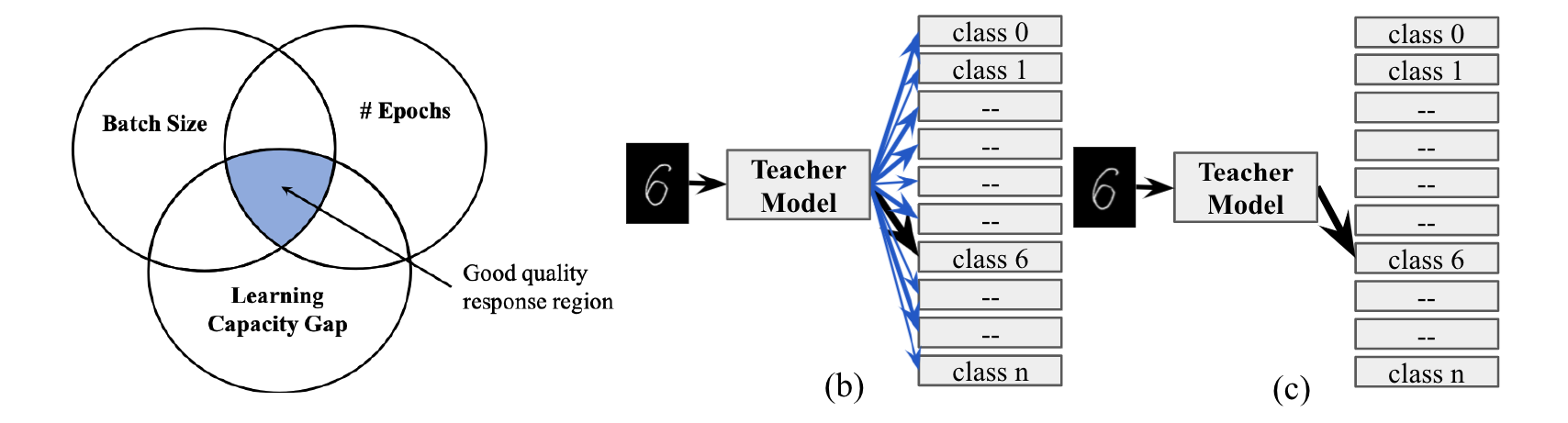}
    \caption{(a) sweet spot between batch size and number of epochs for a given teacher-student pair for effective KD (b) and (c) difference in teacher's response in presence and absence of similarity information, respectively.}
    \label{fig:similarity information relation}
\end{figure}

Wider and deeper deep learning (DL) models have helped us build more and more sophisticated systems across a broad range of areas \cite{lecun_deep_2015}. But the huge computation and memory requirements of these models create hurdles in its deployment on edge devices \cite{simonyan_very_2015, howard_mobilenets_2017}. A number of network compression techniques have been developed to compress large DL models into computationally efficient models with comparable performance, like model quantization \cite{wu_quantized_2016}, model binarization \cite{courbariaux_binaryconnect_2016}, parameter sharing \cite{han_learning_2015, wang_knowledge_2021}, low-rank factorization \cite{denton_exploiting_2014, yu_compressing_2017}, convolution filter compression \cite{zhai_doubly_2016} and knowledge distillation (KD) \cite{bucilua_model_2006, hinton_distilling_2015}. KD has two basic steps, knowledge extraction and distillation. Knowledge extraction can be response-based, feature-based, or relation-based and its distillation can be offline, online, or self-distillation \cite{gou_knowledge_2021}. In this paper, we explore the KD process that uses a teacher's response to distill the knowledge in an offline manner to achieve network compression.


Recent advances suggest that the performance of the response-based offline knowledge distillation (RBKD) process is largely affected by the gap in learning capacities of the teacher-student pair \cite{yuan_revisiting_2020, mirzadeh_improved_2020, gao_residual_2020}. However, we argue that this gap is not the root cause of the poor distillation. 

Our experiments suggest that the quality of distillation is largely controlled by the amount of "similarity information" about classes that a teacher provides in its responses. In other words, teachers provide a distribution over the possible class labels for any given example. A highly-confident teacher produces a "hard" result that has a peak response for exactly one class, whereas a less-confident teacher provides a more spread, higher-entropy distribution. We find, somewhat counter-intuitively, that the less-confident teacher provides the student more information about the relationships between classes through the non-zero responses to the other classes. We find that careful training of the teacher can largely offset the negative effect of the gap between models. Fig. \ref{fig:similarity information relation} (a) presents the idea. For a given teacher-student pair of any capacity gap, we can find a balance between batch size and number of epochs of training a teacher to retain similarity information in its response. Fig. \ref{fig:similarity information relation} (b) and (c) show the fundamental difference in the quality of a teacher's response in the presence and absence of similarity information, respectively. A good response has relevant information about similarity between two classes and a poor response lacks this information. The RBKD process is fast and effective with the similarity-rich response as it learns about more than one class from a single input - \textit{one example-many class} learning. It requires fewer examples per class to learn about the whole distribution of examples. In contrast, the process is slow, less effective, and requires more examples per class when similarity information is absent. 

\begin{figure}[t]
    \centering
    \includegraphics[scale=0.6]{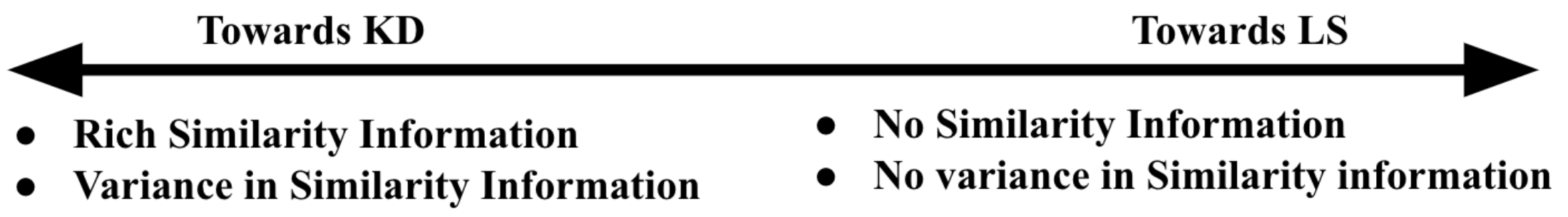}
    \caption{The distillation hypothesis}
    \label{fig:distillation hypothesis text}
\end{figure}

We also observe that for a fixed student network, as the teacher model becomes more powerful, wider and deeper in size, it loses the rich similarity information present in the soft labels and behaves like a label smoothing (LS) \cite{muller_when_2019, szegedy_rethinking_2016} process. We use this observation to differentiate the nature of the distillation process between KD and regularization effect through \textit{the distillation hypothesis} shown in Fig. \ref{fig:distillation hypothesis text}. We argue that any distillation process with little or no similarity information in the teacher's response brings more of a regularization effect than the distillation effect and when it completely lacks similarity information, the KD process is equivalent to the LS.

Our contributions are as follows: (i) We show that rich similarity information in teacher's response can facilitate \textit{one example-many classes} learning and accelerates KD. (ii) We argue that the gap in teacher-student pair is not the root cause or the dominating factor of poor distillation, and provide a method to carefully train any teacher to retain more similarity information in its response to achieve better distillation. (iii) We also propose \textit{the distillation hypothesis} to understand the underlying nature of the distillation process.

\section{Related Work}\label{related work}

\citet{bucilua_model_2006} devise a method of network compression by distilling the knowledge of a teacher network, input to the softmax layer, to a student network. \citet{hinton_distilling_2015} popularized this idea by introducing a temperature term $T$, $q{_i}=\frac{exp(\frac{z_i}{T})}{\sum_{j}exp(\frac{z_j}{T})}$ class probabilities ($q_i$) and logits ($z_i$), in softmax that controls the softness of response. Increasing $T$ decreases $q_i$ for correct class, indicating decrease in the confidence of network's response, while increasing $q_i$ for incorrect classes, indicating increase in similarity information in network's response.

The response-based network compression is a very effective tool to instill the knowledge of a cumbersome model into an efficient model to achieve uncompromised network compression. A common understanding is that a more powerful teacher should be able to provide more knowledge to its student. But it is observed that wider and deeper networks do poor distillation \cite{mirzadeh_improved_2020, yuan_revisiting_2020, gao_residual_2020}. \citet{kim_transferring_2017} and \citet{muller_when_2019} ascribe the effectiveness of RBKD being similar to the effectiveness of LS. \citet{ding_adaptive_2019} explain RBKD as the regularization effect brought about by the response of the teacher model. However, the RBKD process relies on the teacher's response and can not properly explain the hidden-layer supervision \cite{gou_knowledge_2021}. We argue that whether RBKD behaves as regularization, distillation, or LS, depends on the presence of similarity information in the teacher's response. This, in tern, affects the extent of network compression that can be achieved by this method.

\citet{mirzadeh_improved_2020} attribute the gap in learning capacities of teacher-student pair as the only reason for poor distillation, and propose a teacher assistant (TA) model, with a learning capacity in between teacher and student, to address this problem. The knowledge from the teacher is routed through one or more TA(s) to the student. While TAs improve the distillation performance but it also makes the whole process of network compression computationally expensive.

With a similar hypothesis, \citet{yuan_revisiting_2020} propose a teacher-free KD (TFKD) process to reduce the gap between teacher-student pairs. The TFKD process distills its knowledge to itself during training. Theoretically, TFKD reduces the gap to zero, but its knowledge is limited by its learning capacity and extent of training. It also deviates from the network compression task. 

\citet{gao_residual_2020} propose residual KD to distill the knowledge by introducing an assistant model to learn the residual error between teacher and assistant models. This method also tries to reduce the gap between teacher-student pairs. This method is also effective, but it does bring additional computational costs with added networks to achieve network compression.

All the above methods try to find a solution to reduce the gap between the teacher-student pair. They focus specifically on the distillation process and not on the knowledge inculcation process. In this paper, we show that the quality of knowledge distillation can be controlled by controlling the quality of knowledge inculcation in a teacher. We also explain how the nature of KD changes with a change in similarity information in the teacher's response with \textit{the distillation hypothesis}.


\begin{table}[t]
\centering
\renewcommand{\arraystretch}{1}

\begin{tabular}{l l l l l l l l l l l}
\toprule
Input class & \multicolumn{10}{c}{Response of LS process} \\
\cmidrule{2-11}
 & 0 & 1 & 2 & 3 & 4 & 5 & 6 & 7 & 8 & 9 \\
\toprule
\multirow{2}{3em}{digit 6} & 0.06 & 0.06 & 0.06 & 0.06 & 0.06 & 0.06 & \textbf{0.46} & 0.06 & 0.06 & 0.06 \\
& 0.06 & 0.06 & 0.06 & 0.06 & 0.06 & 0.06 & \textbf{0.46} & 0.06 & 0.06 & 0.06 \\
& 0.06 & 0.06 & 0.06 & 0.06 & 0.06 & 0.06 & \textbf{0.46} & 0.06 & 0.06 & 0.06 \\
\bottomrule
\end{tabular}%

\caption{The LS labels for different input images of same class at $\alpha_{LS} = 0.6$. It shows that the distribution of soft-labels in the absence of similarity information. This distribution promotes \textit{one example-one class} learning.}
\label{table:Soft-labels for label smoothing process}
\end{table}

\section{Response Based Techniques} \label{thoery of LS and RBKD}

\subsection{Label Smoothing} \label{label smoothing process}

The LS process uses a less confident form of one-hot label to train a model. It decreases the model confidence about the correct class and treats all other classes as equal, providing no similarity information \cite{muller_when_2019, szegedy_rethinking_2016}. The LS labels are computed as,

\begin{equation}
    \label{eq:label smoothing formula with one-hot vector}
    y_{c}^{LS} = ( 1- \alpha_{LS}) y_{c} + \alpha_{LS} \frac{1}{C}
\end{equation}

where, $y_c$ is the one-hot vector, $c$ is the number of classes, and $1/c$ is a uniform distribution. Examples are shown in Table \ref{table:Soft-labels for label smoothing process}. To calculate the cross-entropy over these labels, we denote the true soft-label distribution $q(c|x)$ ($x$ is input) as $q^{'}$ for LS process, $q$ for RBKD process and use $p$ to denote distribution $p(c|x)$ generated by model \cite{yuan_revisiting_2020}. The loss function is then,

\begin{equation}
\label{Equation: Loss function of LS process}
    \mathcal{L}_{LS} = (1- \alpha_{LS}) H(q,p) + \alpha_{LS} D_{KL}(u,p)
\end{equation}

where $H(u)$ is a fixed entropy value of uniform distribution and $D_{KL}$ is the Kullback-Leibler divergence (KL), 

\begin{equation}
\label{eq:Cross-entropy calculation over LS}
\begin{split}
H(q^{'}, p) &= - \sum_{c=1}^{c} q^{'} log(p)\\
&= (1 - \alpha_{LS})\; H(q, p) + \alpha_{LS} H(u, p)\\
&= (1-\alpha_{LS})\; H(q,p) + \alpha_{LS} (D_{KL}(u,p) \\
&+ H(u))\\
\end{split}
\end{equation}

The soft-labels on the LS process is shown in Table \ref{table:Soft-labels for label smoothing process}. The LS process use a fixed distribution to produce "hard" result that has a peak response for exactly one class, and assigns all other classes equal value without considering the similarity between classes. This paradigm promotes \textit{one example-one class} learning (Fig. \ref{fig:similarity information relation} (c)) i.e. each example provides useful information only about its class and treat all other classes equally.

\subsection{Response-Based Knowledge Distillation} \label{RBKD}

The RBKD process has the teacher produce probabilistic distribution for each class at a temperature $T$. It contains rich similarity information between classes to support \textit{one examples-many classes} learning (Fig. \ref{fig:similarity information relation} (b)) \cite{hinton_distilling_2015}. These information-rich labels are used to distill the knowledge to the student by minimizing the weighted sum of KL divergence and cross-entropy losses as,

\begin{equation}
\label{eq: Loss function of KD process}
    \mathcal{L}_{KD} = (1- \alpha_{KD}) H(p,q) + \alpha_{KD} D_{KL}(p_{T}^{t},p_{T})
\end{equation}

\noindent where $H(p, q)$, $p$, $p_{T}$, and $p_{T}^{t}$ denote the cross-entropy loss for student with true labels ($q$), output of student model, output of student model softened at $T$ and output of teacher model softened at $T$, respectively. $\alpha_{KD}$ is the contribution factor for loss function.

The loss functions of RBKD and LS processes have similar formulations as shown in Equation (\ref{Equation: Loss function of LS process}) and (\ref{eq: Loss function of KD process}), but they use different methods for generating soft-labels. The only difference is the $p_{T}^{t}$ in $D_{KL}(p_{T}^{t},p_{T})$, which is generated by a teacher model, and $u$ in $D_{KL}(u,p)$, which is a uniform distribution \cite{yuan_revisiting_2020}. This difference decides the extent of similarity information in the response, and that governs the nature of the distillation process. We discuss this later in Section \ref{Experiments and Results}. It is safe to conclude that LS is a special case of the RBKD process in which soft labels are generated by a constant distribution as prior knowledge instead of learned-knowledge of a pre-trained teacher.

\begin{table}[t]
\centering
\renewcommand{\arraystretch}{1}

\begin{tabular}{l l l l l l l l l l l}
\toprule
Input class & \multicolumn{10}{c}{Soft-labels by small learning capacity teacher} \\
\cmidrule{2-11}
 & 0 & 1 & 2 & 3 & 4 & 5 & 6 & 7 & 8 & 9 \\
\toprule
\multirow{5}{3em}{Digit 6} & 0.087 & 0.048 & 0.095 & 0.093 & 0.077 & 0.144 & \textbf{0.206} & 0.051 & 0.122 & 0.078 \\
& 0.087 & 0.048 & 0.089 & 0.1 & 0.090 & 0.119 & \textbf{0.177}  & 0.056 & 0.114 & 0.095 \\
& 0.090 & 0.078 & 0.089 & 0.097 & 0.115 & 0.091 & \textbf{0.179} & 0.071 & 0.108 & 0.082 \\
& 0.107 & 0.068 & 0.089 & 0.076 & 0.104 & 0.1 & \textbf{0.229} & 0.070 & 0.086 & 0.071 \\
& 0.118 & 0.079 & 0.095 & 0.075 & 0.101 & 0.081 & \textbf{0.210} & 0.069 & 0.098 & 0.073 \\
\bottomrule
\end{tabular}%

\caption{Soft-labels generated by small capacity teacher model with 3 hidden layers on MNIST data.}
\label{Table: Soft-labels by small learning capacity teacher}

\renewcommand{\arraystretch}{1}

\begin{tabular}{l l l l l l l l l l l}
\toprule
Input class & \multicolumn{10}{c}{Soft-labels by large learning capacity teacher} \\
\cmidrule{2-11}
 & 0 & 1 & 2 & 3 & 4 & 5 & 6 & 7 & 8 & 9 \\
\toprule
\multirow{5}{3em}{Digit 6} & 0.078 & 0.069 & 0.063 & 0.059 & 0.070 & 0.135 & \textbf{0.356} & 0.039 & 0.078 & 0.053 \\
& 0.083 & 0.077 & 0.073 & 0.056 & 0.078 & 0.107 & \textbf{0.339}  & 0.042 & 0.090 & 0.053 \\
& 0.077 & 0.067 & 0.068 & 0.047 & 0.073 & 0.086 & \textbf{0.425} & 0.034 & 0.079 & 0.044 \\
& 0.076 & 0.062 & 0.057 & 0.039 & 0.059 & 0.090 & \textbf{0.485} & 0.027 & 0.065 & 0.039 \\
& 0.077 & 0.067 & 0.061 & 0.043 & 0.068 & 0.089 & \textbf{0.450} & 0.031 & 0.075 & 0.041 \\
\bottomrule
\end{tabular}%

\caption{Soft-labels generated by a large capacity teacher model with 6 hidden layers on MNIST data}
\label{Table: Soft-labels generated by a large learning capacity teacher model}
\end{table}


\subsection{Quality of Teacher's Response} \label{quality of teacher's response}

The student mimics the teacher's response by minimizing the KL divergence between their response at $T$. The RBKD loss, Equation (\ref{eq: Loss function of KD process}), is weighed higher $\alpha_{KD}$, $0.99$, for KD. Each example generates two types of soft-labels, for itself - \textit{confidence label}, and for all other classes - \textit{similarity labels}. The confidence label shows the confidence of a teacher for the correct class and the similarity labels provide a probabilistic value of similarity of all other classes with input class. Tables \ref{table:Soft-labels for label smoothing process}, \ref{Table: Soft-labels by small learning capacity teacher}, and \ref{Table: Soft-labels generated by a large learning capacity teacher model} show the confidence label in bold text and similarity labels in plain text. The hyper-parameter $\alpha_{LS}$ is $0.6$ for LS process in Table \ref{table:Soft-labels for label smoothing process}, while temperature $T$ is $9$ for RBKD process in Tables \ref{Table: Soft-labels by small learning capacity teacher} and \ref{Table: Soft-labels generated by a large learning capacity teacher model}. 

The quality of distillation is controlled by the similarity information in the teacher's response, which can be controlled through $T$. But the increase and decrease in similarity information are not linear with $T$, but instead dependent on the similarity relation learned by the teacher during knowledge inculcation. The teacher perceives each example differently, even within the same class, and assigns a different confidence and similarity labels in its response. This introduces a variation in confidence and similarity labels ---\textit{a variance in response}, as shown row-wise and column-wise in Tables \ref{Table: Soft-labels by small learning capacity teacher} and \ref{Table: Soft-labels generated by a large learning capacity teacher model}. More variance in similarity is desirable for KD as it provides knowledge about which classes are most similar to others. A highly confident teacher produces less variance as compared to a less confident teacher in its response, degrading the quality of distillation. 

The behavior of a highly confident teacher model is analogous to the LS process. The similarity labels are treated equally with a little or no variance in response, providing neither similarity information nor variance in response. We describe this observation through the distillation hypothesis.


\begin{figure}[t]
    \centering
    \includegraphics[width=0.98\columnwidth]{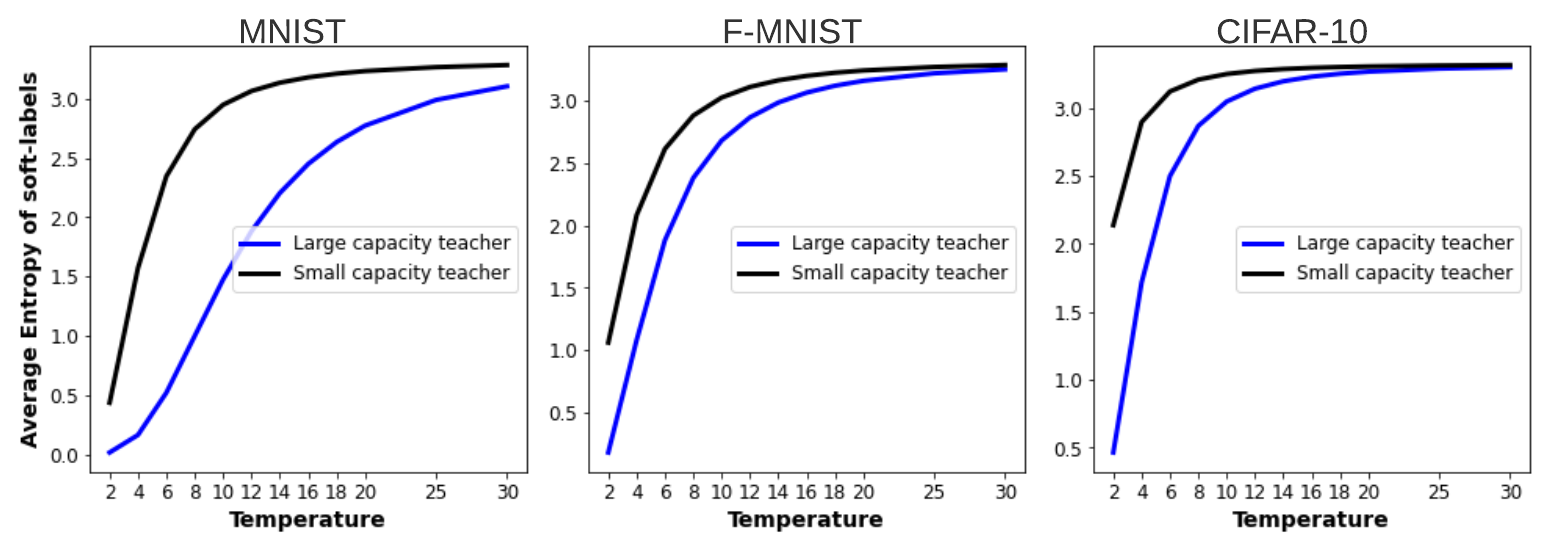}
    \caption{Average entropy of teacher's response for MNIST, Fashion-MNIST and CIFAR-10 datasets.}
    \label{fig:average entorpy of soft labels}
\end{figure}

\begin{table}[t]
\centering
\renewcommand{\arraystretch}{0.8}

\begin{tabular}{l l l l l l l l l l l l l l l}
\toprule
\textbf{Missing} & \multicolumn{7}{c}{\textbf{KD by small capacity teacher}} & \multicolumn{7}{c}{\textbf{KD by large capacity teacher}}\\
\cmidrule{2-15}
\textbf{class} & \multicolumn{7}{c}{Student accuracy(\%) on removed class} & \multicolumn{7}{c}{Student accuracy(\%) on missing class}\\
 & T=3 & T=6 & T=9 & T=12 & T=15 & T=20 & T=3 & T=6 & T=9 & T=12 & T=15 & T=20 \\
\toprule
\textbf{Digit 0} & 99.18 & 99.28 & 99.48 & 99.18 & 98.67 & 98.57 & 2.8 & 0.20 & 0.61 & 0.0 & 0.0 & 0.0 \\
\textbf{Digit 1} & 98.67 & 98.67 & 98.67 & 98.59 & 98.50 & 98.23 & 0.0 & 0.26 & 0.0 & 0.26 & 0.08 & 0.0 \\
\textbf{Digit 2} & 95.73 & 96.80 & 96.12 & 95.83 & 94.86 & 94.76 & 2.5 & 0.38 & 0.58 & 0.48 & 1.25 & 0.19\\
\textbf{Digit 3} & 98.11 & 98.41 & 98.31 & 98.11 & 98.01 & 97.92 & 0.0 & 0.79 & 1.38 & 6.23 & 2.47 & 2.27 \\
\textbf{Digit 4} & 98.47 & 98.67 & 98.57 & 98.67 & 98.37 & 96.94 & 0.10 & 0.10 & 0.0 & 0.0 & 0.0 & 0.0 \\
\textbf{Digit 5} & 96.86 & 97.53 & 97.30 & 97.19 & 96.41 & 97.86 & 0.67 & 0.56 & 1.35 & 2.80 & 2.46 & 4.26 \\
\textbf{Digit 6} & 97.39 & 97.49 & 97.28 & 97.18 & 97.18 & 96.65 & 17.32 & 2.08 & 12.83 & 7.09 & 18.58 & 7.41 \\
\textbf{Digit 7} & 96.10 & 96.30 & 96.40 & 95.91 & 94.84 & 95.03 & 0.29 & 0.29 & 3.50 & 0.58 & 0.77 & 1.84 \\
\textbf{Digit 8} & 97.12 & 96.71 & 97.22 & 96.61 & 96.40 & 96.09 & 0.00 & 0.10 & 0.0 & 0.0 & 0.0 & 0.0 \\
\textbf{Digit 9} & 96.13 & 96.63 & 96.23 & 95.83 & 95.63 & 94.25 & 0.00 & 0.69 & 0.10 & 1.48 & 1.28 & 2.18 \\
\bottomrule

\end{tabular}%

\caption{Student accuracy on missing class (MNIST) when distilled with small and large learning capacity teacher}
\label{table:MNIST: Similarity Information}
\end{table}

\subsection{The Distillation Hypothesis} \label{the distillation hypothesis}

The distillation hypothesis, Figure \ref{fig:distillation hypothesis text}, defines the nature of distillation by observing the quality of similarity labels and variance in response of a teacher. It states that for a given student network, as the learning capacity of the teacher network increases, the nature of the distillation process starts to move away from a similarity label-based RBKD process to a non-similarity-based LS process. The shift towards the LS process is caused by the loss of similarity information and variance in response of a teacher and leads to poor distillation. In other words, the nature of the knowledge distillation process shifts away from \textit{one example-many classes} learning to \textit{one example-one class} learning.

Since the quality of distillation is directly dependent on the knowledge inculcation process of a teacher. We argue that any teacher model can be trained to retain more similarity information by finding a sweet spot between the batch size and the number of epochs for a given teacher-student pair (see Figure \ref{fig:similarity information relation} (a)). The blue region symbolizes the right balance between the batch size and the number of epochs for better knowledge distillation. 

Every DL model has an optimal batch size for optimum learning. The optimal batch size is the number of examples that the model is effectively able to process. If the batch size is larger than its optimal value, then the model is not able to process all the information at once and takes more epochs to achieve high confidence. These two factors can be iteratively balanced to reach the optimal batch size and number of epochs for knowledge inculcation of a teacher.

\section{Experiments and Results} \label{Experiments and Results}

First, we present empirical results to support the distillation hypothesis, and then show the results of improved distillation achieved using our proposed method on MNIST, Fashion-MNIST, and CIFAR-10 datasets.

\begin{table}[t]
\centering
\renewcommand{\arraystretch}{0.8}
\resizebox{\columnwidth}{!}{%
\begin{tabular}{l l l l l l l l l l l l l l l}
\toprule
\textbf{Missing} & \multicolumn{7}{c}{\textbf{KD by small capacity teacher}} & \multicolumn{7}{c}{\textbf{KD by large capacity teacher}}\\
\cmidrule{2-15}
\textbf{class} & \multicolumn{7}{c}{Student accuracy(\%) on removed class} & \multicolumn{7}{c}{Student accuracy(\%) on missing class}\\
 & T=3 & T=6 & T=9 & T=12 & T=15 & T=20 & T=3 & T=6 & T=9 & T=12 & T=15 & T=20 \\
\toprule
\textbf{T-shirt}  & 84.70 & 77.20 & 62.8 & 46.70 & 32.69  & 23.70 & 0.6 & 6 & 7.6 & 7 & 8.4 & 9\\
\textbf{Trouser} & 94.90 & 94.19 & 92.90 & 91.39 & 88.09  & 89.99 & 0.0 & 4.2 & 15.4 & 19.2 & 23.2 & 16.8\\
\textbf{Pullover} & 64.89 & 53.79 & 44.20 & 28.99 & 17.49 & 15.00 & 0.0 & 0.0 & 0.1 & 0.1 & 0.1 & 0.1 \\
\textbf{Dress} & 84.50 & 82.09 & 76.70 & 69.99 & 60.50 & 46.90 & 0.8 & 0.7 & 1.8 & 1.0 & 1.7 & 1.4\\
\textbf{Coat} & 79.19 & 71.49 & 59.79 & 44.49 & 28.40 & 16.20 & 0.3 & 0.4 & 0.4 & 1.4 & 1.0  & 1.6\\
\textbf{Sandal} & 91.50 & 92.00 & 90.49 & 87.30 & 84.89 & 82.30 & 0.2 & 0.4 & 0.4 & 0.4 & 0.4 & 0.5\\
\textbf{Shirt} & 55.80 & 42.30 & 30.30 & 17.49 & 6.40 & 2.70 & 0.2 & 0.7 & 1.1 & 1.7 & 2.1  & 1.9\\
\textbf{Sneaker} & 91.79 & 87.69 & 83.30 & 69.99 & 63.89 & 57.09 & 0.1 & 0.0 & 0.0 & 0.0 & 0.0 & 0.0\\
\textbf{Bag} & 95.20 & 95.20 & 94.19 & 91.60 & 88.80 & 86.19 & 0.0 & 0.5 & 0.4 & 0.4 & 0.9 & 0.4\\
\textbf{boot} & 91.29 & 89.99 & 87.99 & 83.39 & 79.69 & 74.00 & 1.1 & 2.6 & 4.4 & 3.6 & 3.7 & 3.7 \\
\bottomrule
\end{tabular}%
}
\caption{Student accuracy on missing class (Fashion-MNIST) when distilled with small and large learning capacity teacher}
\label{table: F-MNIST: Similarity Information}


\renewcommand{\arraystretch}{0.8}

\begin{tabular}{l l l l l l l l l l l l l l l}
\toprule
\textbf{Missing} & \multicolumn{7}{c}{\textbf{KD by small capacity teacher}} & \multicolumn{7}{c}{\textbf{KD by large capacity teacher}}\\
\cmidrule{2-15}
\textbf{class} & \multicolumn{7}{c}{Student accuracy(\%) on removed class} & \multicolumn{7}{c}{Student accuracy(\%) on missing class}\\
 & T=3 & T=6 & T=9 & T=12 & T=15 & T=20 & T=3 & T=6 & T=9 & T=12 & T=15 & T=20 \\
\toprule
\textbf{plane} & 70.99 & 66.6 & 52.1 & 42.6 & 39.7 & 31.5 & 1.79 & 0.89 & 0.49 & 0.70 & 0.80 & 0.60 \\
\textbf{Auto} & 80.8 & 77.5 & 70.6 & 63.8 & 55.8 & 45 & 2.09 & 0.70 & 0.30 & 0.20 & 0.30 & 0.10\\
\textbf{Bird} & 48.1 & 41.8 & 35 & 26.9 & 25.7 & 14 & 0.89 & 0.60 & 0.30 & 0.40 & 0.30 & 0.20 \\
\textbf{Cat} & 50.8 & 42.1 & 33 & 22.6 & 8.7 & 4.6 & 0.80 & 0.70 & 0.49 & 0.30 & 0.30  & 0.10\\
\textbf{Deer} & 61.59 & 53.2 & 55.7 & 43.1 & 26.2 & 16.2 & 0.99 & 1.49 & 0.89 & 0.60 & 0.80 & 0.40\\
\textbf{Dog} & 67.4 & 55.6 & 29.1 & 20.2 & 16 & 10.8 & 0.70 & 0.70 & 0.40 & 0.20 & 0.30 & 0.20 \\
\textbf{Frog} & 79.5 & 76.3 & 65.9 & 60.5 & 47.1 & 37.7 & 2.30 & 1.89 & 1.60 & 1.99 & 1.4 & 2.09\\
\textbf{Horse} & 67.69 & 63.8 & 63.2 & 54.2 & 52.7 & 40 & 0.80 & 0.20 & 0.40 & 0.40 & 0.20  & 0.49\\
\textbf{Ship} & 73.79 & 70.3 & 67.5 & 59.7 & 54 & 46.3 & 3.99 & 2.09 & 1.49 & 1.09 & 1.09 & 0.49\\
\textbf{Truck} & 79.4 & 75.4 & 73.3 & 66.8 & 55.1 & 43.3 & 3.09 & 2.99 & 1.09 & 0.80 & 0.09 & 0.80\\
\bottomrule

\end{tabular}%

\caption{Student accuracy on missing class (CIFAR 10) when distilled with small and large learning capacity teacher}
\label{table: CIFAR 10: Similarity Information}
\end{table}

\subsection{Similarity Information in Teacher's Response}

The entropy of the soft labels is directly proportional to the presence of similarity information in the teacher's response. It is calculated as $E_{Soft-labels} = - \sum_{i=1}^{C} p_i log(p_i)$  where $C$ is the number of classes in the dataset, $p_i$ is the probabilistic value from teacher. We use two different capacities of teachers, one with large learning capacity and the second with small learning capacity, to study this behavior. We argue that small capacity teachers having fewer parameters are not able to learn the fine-grained discriminative properties between classes and retain more similarity information in their response as compared to the large capacity teacher which can learn fine-grained properties and lose similarity information in the process. 

We show two experiments to support our argument. First, we compare the average entropy in teachers' responses for both small and large capacity teachers at different $T$, see Figure \ref{fig:average entorpy of soft labels}. At all $T$ the average entropy of small capacity teachers remains higher than large capacity teachers. In other words, it is safe to say that a moderately confused teacher is better for knowledge distillation as compared to a well-trained large capacity teacher.

Second, we scrutinize the impact of similarity information for KD. We remove one out of $N$ classes from distillation set during the KD process, and then check the performance of the student on the removed class, see Tables \ref{table:MNIST: Similarity Information}, \ref{table: F-MNIST: Similarity Information}, and \ref{table: CIFAR 10: Similarity Information}. For the student network, the missing class is something it has never seen during training. It learns about the removed class only through the probabilistic value of the similarity information in the teacher's response. 

\begin{figure}[t]
    \centering
    \includegraphics[width=0.98\columnwidth]{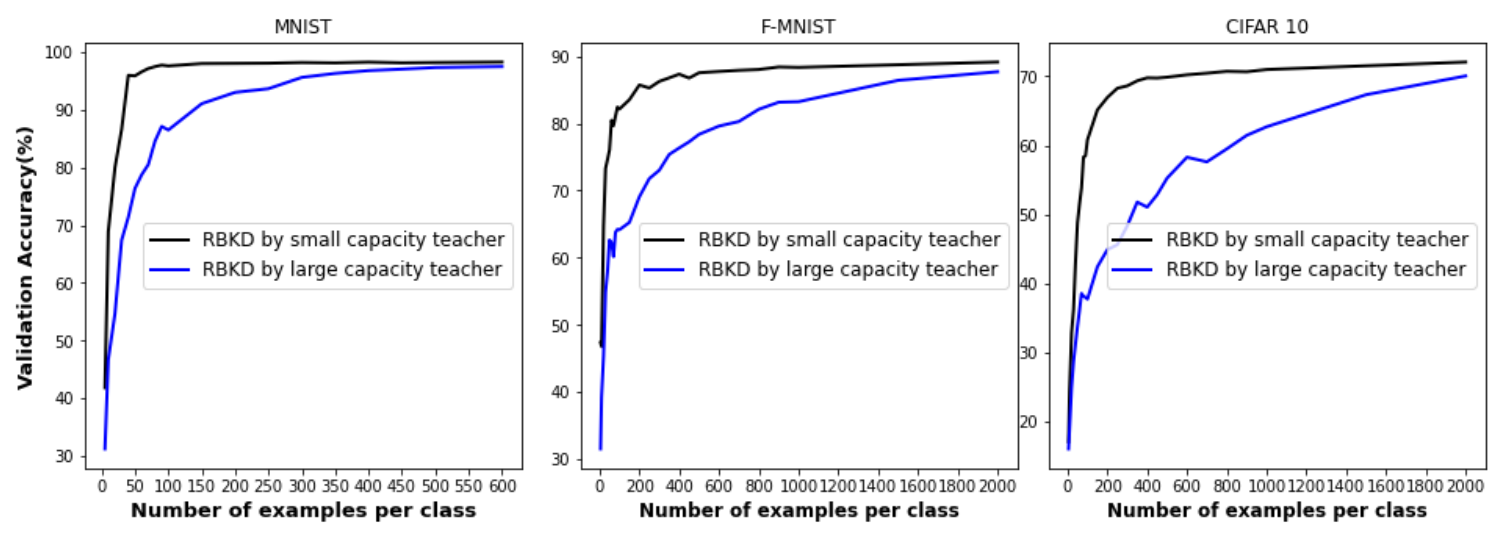}
    \caption{Entropy based example selection for knowledge distillation. RBKD process is more efficient if it requires lesser number of examples per class to distill the knowledge.}
    \label{fig:entropy based transferset selection}
    \includegraphics[width=0.98\columnwidth]{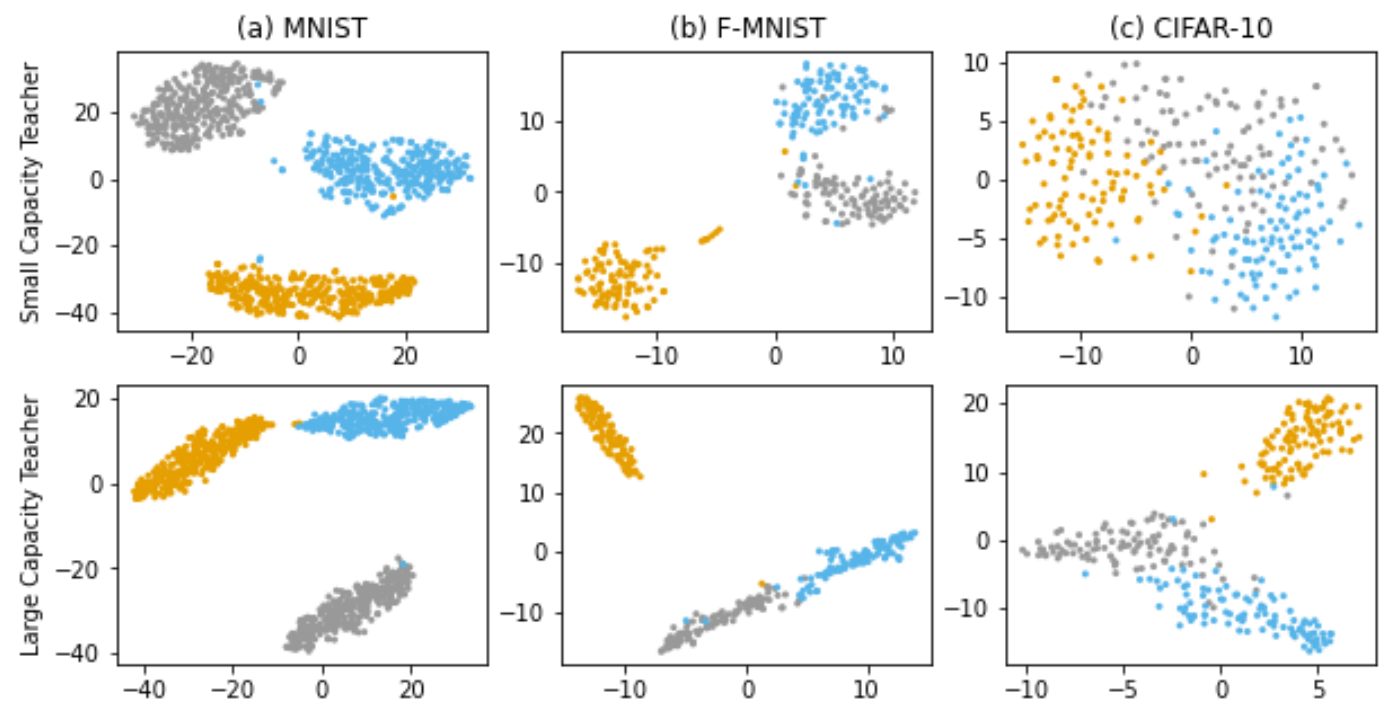}
    \caption{Penultimate layer representation of small and large capacity teacher models.}
    \label{fig:penultimate layer representation}
\end{figure}

The impact of similarity information on the quality of distillation is visible in Table \ref{table:MNIST: Similarity Information}, \ref{table: F-MNIST: Similarity Information} and \ref{table: CIFAR 10: Similarity Information}. Each table is divided into two columns, KD by the small and large capacity teachers. For each teacher, we perform KD on the same student and check its accuracy on the removed class on all three datasets. KD by the small teacher is very effective throughout but KD by the large teacher significantly under-performs on the removed class test. This shows that the quality of KD is controlled by the presence of similarity information in the teacher's response. In presence of similarity-rich soft responses to classes, the student can learn about more than one class from a single input example (one example- many class learning), while it can learn about only one class from one input example if the response is "hard". We further argue that this critical information in soft labels defines the nature of distillation to be KD, regularization, or LS. We use entropy as an indicator of similarity information in the rest of the paper.

\subsection{KD is one example-many class learning}

We established that similarity labels are the most critical information for KD. Now, we show that similarity information accelerates the KD process and only requires a handful of examples to distill knowledge from the teacher. With two different capacity teacher models and using entropy as an indicator at $T=9$, we select different number of examples varying from $5$ to $2000$ per class for KD. We compare the highest validation accuracy achieved by the student network at each step as shown in Figure \ref{fig:entropy based transferset selection}. The validation accuracy of the student network distilled by a large capacity teacher always remains smaller than by a small capacity teacher on each dataset. 

For MNIST, distillation by a large capacity teacher requires 10 times more examples per class to achieve similar accuracy, and this gap increases for more complex datasets (15 times for Fashion-MNIST and 18 times for CIFAR-10). This shows that a large capacity teacher with less similarity information in its response fails to distill its knowledge using one example-many class learning and requires a much larger transfer set to achieve the same performance. This also establishes that one example-many class learning is a very important component of a good KD process.

\begin{figure}[t]
    \centering
    \includegraphics[width=0.98\columnwidth]{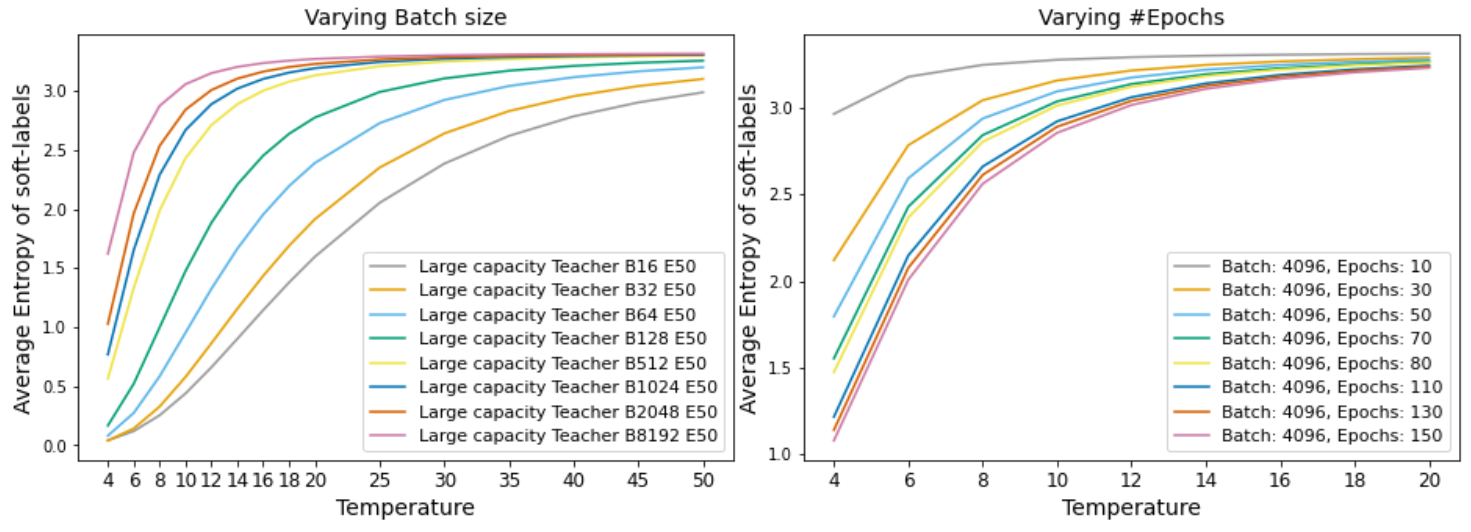}
    \caption{Variation in average entropy of soft-label vs temperature. Left, for different batch sizes at epoch 50. Right, for different epochs at batch size 4096 on MNIST dataset.}
    \label{fig:fixed batch epoch variation}
    \centering
    \includegraphics[width=0.98\columnwidth]{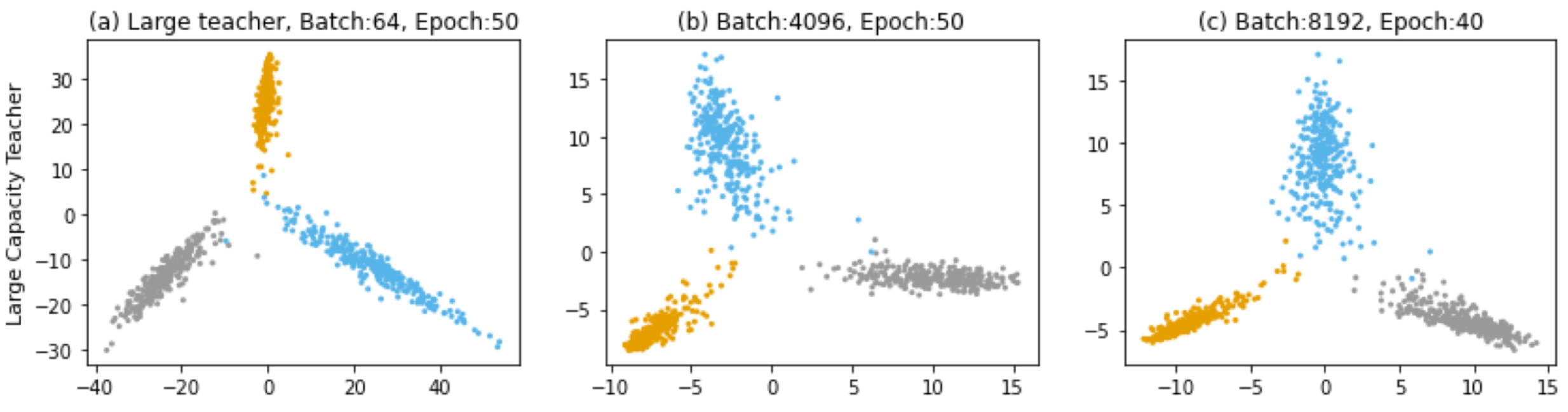}
    \caption{Improved penultimate layer representation for large capacity teach on MNIST. (a) shows the tight clusters, (b) and (c) shows similar clustering for different batch sizes and number of epochs.}
    \label{fig:improved layer representation}
\end{figure}

\subsection{Penultimate Layer Representation}

\citet{muller_when_2019} presents a visualization technique to understand the effects of the LS process on penultimate layer representations of a model. This method uses the linear projection of the activation of the penultimate layer to understand the change in representation. \citet{muller_when_2019} argue that clusters that are relatively spread carry more similarity information in their response as compared to compact clusters. 

The distillation hypothesis says that with the increase in learning capacity of a teacher model it loses similarity information in its response. This loss of similarity indicates that the nature of distillation shifts towards the nature of the LS process. We use this experiment to corroborate the distillation hypothesis by showing the change in clustering of penultimate layer projection which directly relates to similarity information in soft labels.

We randomly select three classes and then choose 300 examples from each class randomly. We extract its activation from the penultimate layer of both small and large capacity models and plot its projection, see Figure \ref{fig:penultimate layer representation}. We observe that the penultimate layer representation of the stronger teacher model forms tighter clusters as compared to small capacity teachers for each dataset. The spread-out clusters indicate that the presence of similarity information between classes that transcends to soft-labels generated by respective teachers, whereas the tighter cluster indicates the absence of similarity information between classes. This provides strong evidence for the distillation hypothesis.

\section{Similarity-Rich Knowledge Inculcation}

\begin{figure}[t]
    \centering
    \includegraphics[scale=0.7]{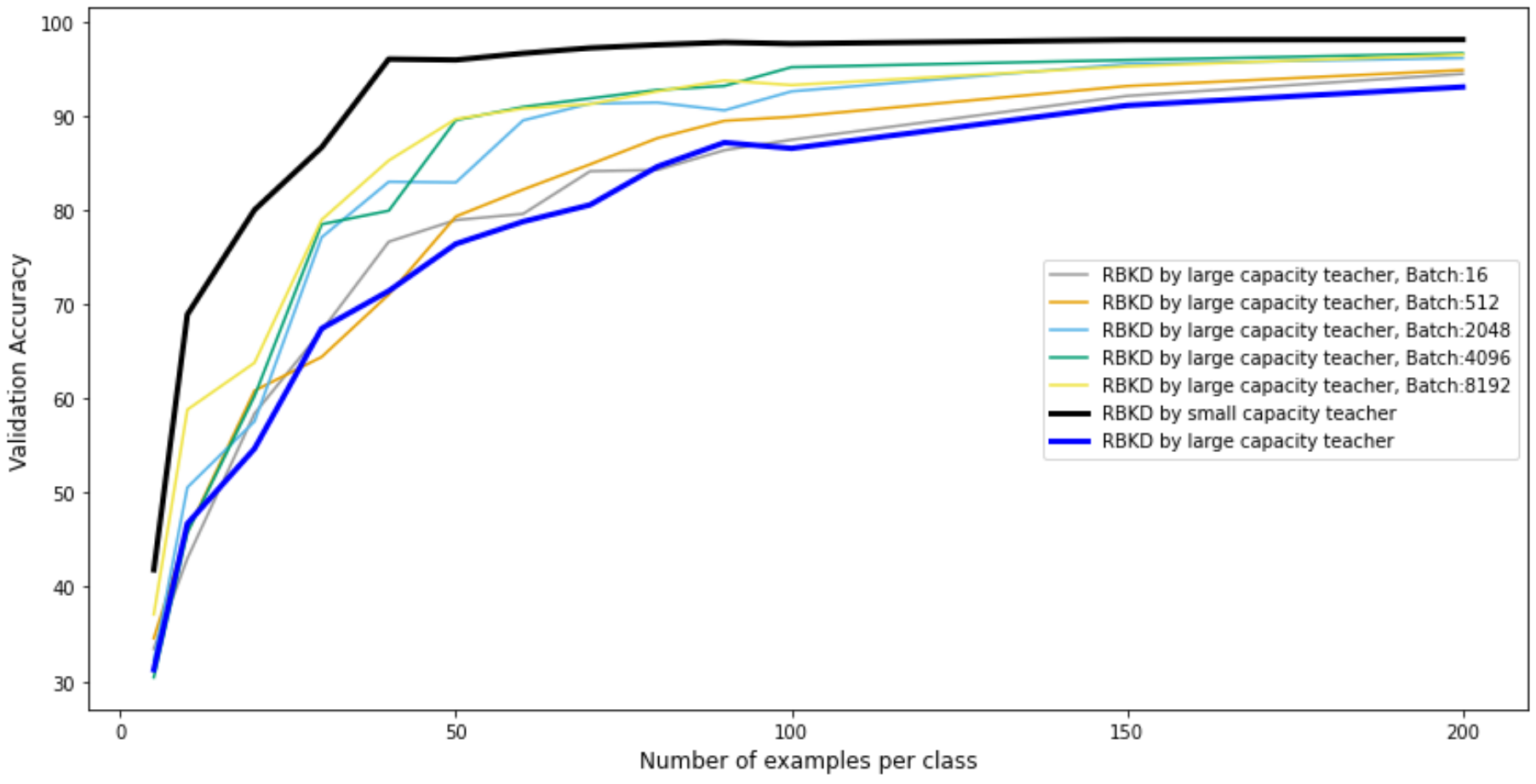}
    \caption{Improved number of examples required per class for distillation on MNIST}
    \label{fig:improved number of examples}
\end{figure}

We already established the importance of similarity information in a teacher's response. But instead of trying to improve the KD process, we focus on the knowledge inculcation process - training the teacher. We carefully calibrate the optimum value of batch size and number of epochs to train the teacher described in section \ref{the distillation hypothesis}. We show improvement for each experiment shown before.

Based on our hypothesis, we show the change in average entropy of teachers' responses for different batch sizes and number of epochs in Figure \ref{fig:fixed batch epoch variation}. It shows that to increase the entropy of soft labels, we can either increase in batch size at a fixed epoch or we can decrease the number of epochs for given batch size. By carefully and iteratively balancing these two factors, we can find one of the many possible sweet spots suited for KD. This controls the extent of similarity information in the teacher's response, which is responsible for the quality of KD and network compression in this process.  

Next, we show the improvement in penultimate layer representation of the large capacity teacher by adapting to our method of knowledge inculcation in Figure \ref{fig:improved layer representation}. Figure \ref{fig:improved layer representation}(a) shows tight clusters, indicating a lack of similarity information. These clusters spread out for many combinations of batch sizes and number of epochs indicating more similarity information in the teacher’s response. A similar clustering is observed for batch size 4096 - epochs 50 and for batch size 8192 - epochs 40 in Fig. \ref{fig:improved layer representation}(b) and (c), respectively. We can conclude that there can be many combinations of batch size and number of epochs for a given teacher-student pair for a more efficient KD.

With all these improvements in the quality of teachers' responses, the minimum number of examples needed to achieve optimal performance should also reduce for large teachers. As shown in Fig. \ref{fig:improved number of examples}, we improve around 5 times in the minimum number of examples required per class to achieve similar performance, decreased from 500 to 100 examples per class, for efficient KD. This provides further support to our idea of focusing on the knowledge inculcation process for achieving better and faster distillation, leading to better network compression.

\begin{figure}[t]
    \centering
    \includegraphics[scale=0.7]{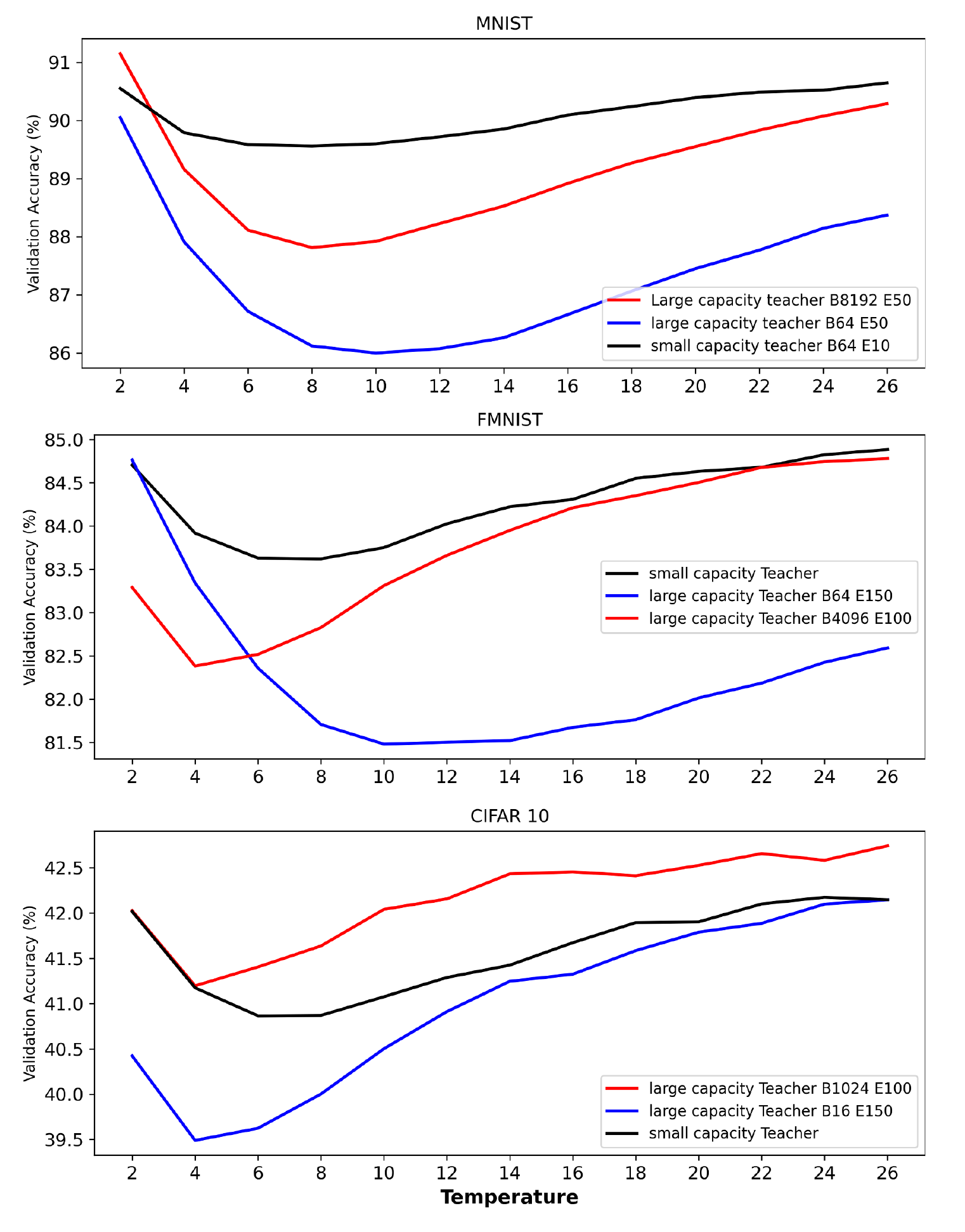}
    \caption{Improved performance on training with suitable batch size and epochs,}
    \label{fig:single layer experiment}
\end{figure}

Even with the above empirical results supporting our hypothesis, we still wanted to test our hypothesis for extreme teacher-student pair situations by maximizing the learning capacity gap between them. To achieve this, we use a single-layer student (baseline student) model with only a classification layer (we provide detail of models used in this paper in Table \ref{table: students model details for all datasets}). We retrain the same large capacity teacher models for each dataset but train it differently - with different batch sizes and number of epochs, to accommodate for the larger gap in model capacities and use it for KD to the baseline student. We also use the small capacity teacher model to show the relative improvements in Fig. \ref{fig:single layer experiment}. The black, blue, and red lines in Fig. \ref{fig:single layer experiment} shows the performance of baseline students on RBKD by small, and large capacity teachers trained with different batch size and epochs at different temperatures. The gap between the red and the blue line shows the improvement in KD performance by a single-layer student.


\begin{table}[t]
\centering
\renewcommand{\arraystretch}{1}
\begin{tabular}{l l l l l l l}
\toprule
Dataset & Capacity & Conv & Dense  & Dropout & BN & Total Params \\
\toprule
MNIST & Large & 3 & 3 & Yes & $\times$ & 2,560,906 \\
 & Small & 2 & 1 & $\times$ & $\times$ & 1,433,610 \\
F-MNIST & Large & 4 & 4 & Yes & $\times$ & 2,339,850 \\
 & Small & 3 & 1 & $\times$ & $\times$ & 1,558,538 \\
CIFAR-10 & Large & 8 & 3 & Yes & Yes & 26,902,442 \\
 & Small & 3 & 1 & $\times$ & $\times$ & 5,674,634 \\
\bottomrule

Dataset & Student Type & Conv & Dense  & Dropout & BN & Total Params \\
\toprule
MNIST & General & 2 & 1 &  $\times$ & $\times$ & 20,490 \\
 & Baseline & $\times$ & 1 &  $\times$ & $\times$ & 7,850 \\
F-MNIST & General & 2 & 1 &  $\times$ & $\times$ & 50,186 \\
 & Baseline & $\times$ & 1 &  $\times$ & $\times$ & 7,850 \\
CIFAR-10 & General & 3 & 1 &  $\times$ &  $\times$ & 534,666 \\
 & Baseline & $\times$ & 1 &  $\times$ &  $\times$ & 30,730 \\
\bottomrule
\end{tabular}%

\caption{Details of teacher and student models for different datasets. Conv, Dense, Dropout, and BN denote the number of convolution layers, number of Dense layers, dropout layers, and Batch Normalization layers, respectively.}
\label{table: students model details for all datasets}
\end{table}

\section{Summary and Conclusion} \label{summary}

\begin{figure}[h!]
    \centering
    \includegraphics[scale=0.6]{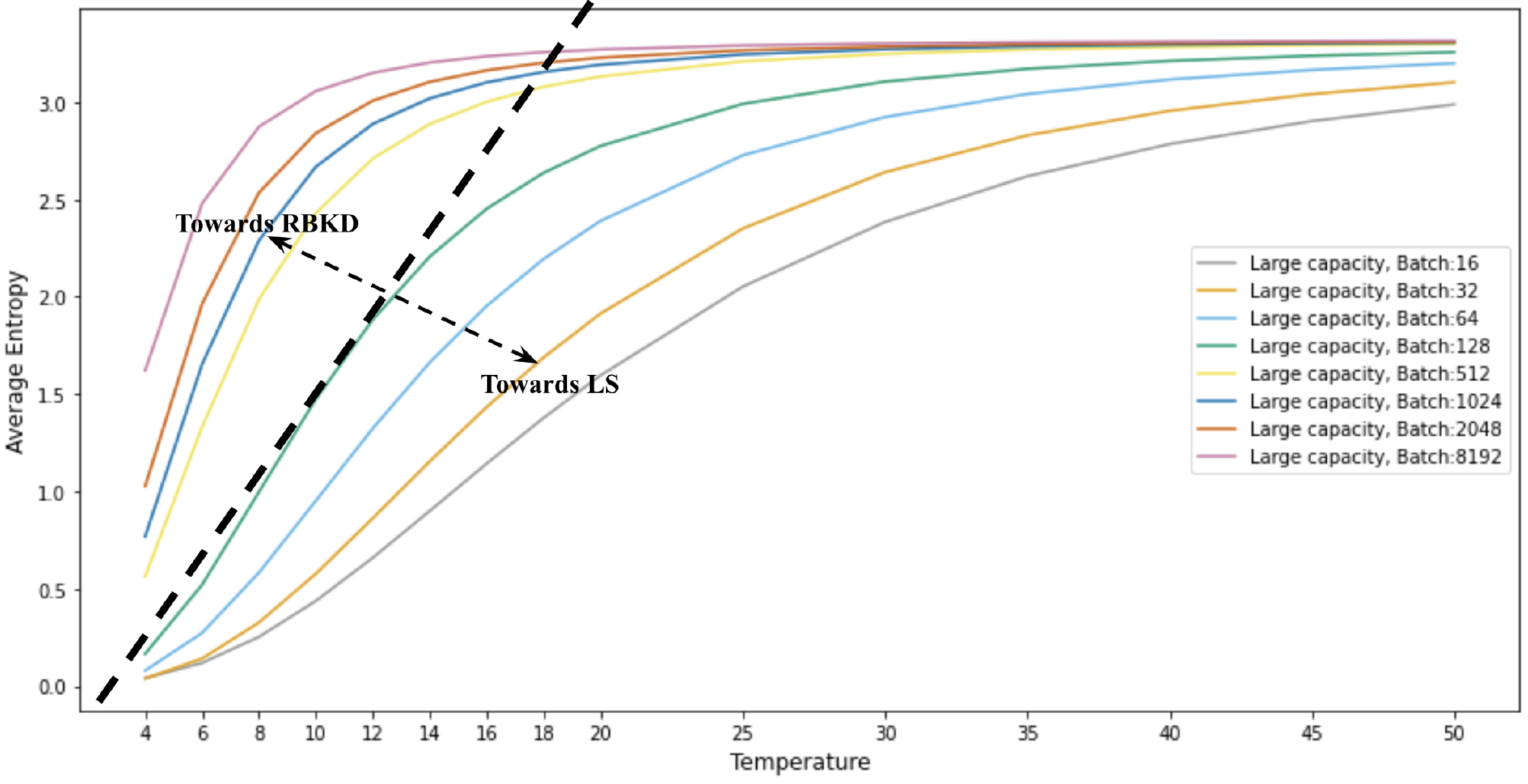}
    \caption{The distillation hypothesis in terms of average entropy of soft-labels.}
    \label{fig:improved average entropy}
\end{figure}

For building intuition about the nature of distillation, we present the entropy-version of the distillation hypothesis (previously shown in Fig. \ref{fig:distillation hypothesis text}) to approximate the nature of the distillation process in Figure \ref{fig:improved average entropy}. For all the processes lying on the extreme ends in Fig. \ref{fig:distillation hypothesis text}, we can imaging multiple lines with different entropy spread across, as shown in Fig. \ref{fig:improved average entropy}. We can draw an imaginary line (the dashed line in the figure) that separates the RBKD process and the LS process. For efficient distillation from any teacher to any student, the corresponding entropy line should lie in the RBKD region. This can be achieved using our proposed method of knowledge inculcation during teacher's training. We emphasize that this representation is to develop an intuitive understanding of the KD process.

We discuss that the quality of distillation can be controlled at various stages of the KD process. While most works focuses on perfecting the distillation step of the KD process, we focus on controlling the quality of distillation by improving the knowledge inculcation process of the teacher model. Our experiments suggest that similarity information in a teacher's response plays a dictating role in determining the quality of distillation. We show the role of similarity information in accelerating the distillation process through one example-many class learning, making the KD process more effective and efficient. The distillation hypothesis can help in approximating the nature of the KD process to be RBKD, regularization, or LS process and to help make the process more effective. While we argue that the distillation performance of any teacher-student pair can be improved by our method of knowledge inculcation, we do not explore the extent to which the KD process can be improved. We believe that the factor of improvement should be different for each teacher-student pair. We leave this for future exploration.

\bibliographystyle{unsrtnat}
\bibliography{distillation}  

\end{document}